\documentclass[11pt]{article}

\usepackage[preprint]{acl}

\usepackage{times}
\usepackage{latexsym}
\usepackage[T1]{fontenc}
\usepackage[utf8]{inputenc}
\usepackage{microtype}

\usepackage{graphicx}
\usepackage{booktabs}
\usepackage{amsmath,amssymb}
\usepackage{xcolor}
\usepackage{colortbl}
\usepackage{array}
\usepackage{placeins}
\usepackage{float}
\usepackage{adjustbox}
\usepackage{tikz}
\usepackage{pifont}
\usepackage{pdfrender}
\usepackage{multirow}
\usepackage{enumitem}
\usetikzlibrary{arrows.meta}
\usepackage{pgfplots}
\pgfplotsset{compat=1.18}
\usepgfplotslibrary{groupplots}

\definecolor{navy}{HTML}{1E3354} %
\definecolor{sky}{HTML}{87B0CF}
\definecolor{skypanel}{HTML}{E7EDF5}
\definecolor{gold}{HTML}{E2BF5C}
\definecolor{goldtext}{HTML}{B98A1C}
\definecolor{accent}{HTML}{2E5E9E} %
\definecolor{accentrule}{HTML}{A9B9CC} %
\definecolor{oursbg}{HTML}{FDF5E2} %
\definecolor{bandbg}{HTML}{F1F5FA}
\definecolor{panelbg}{HTML}{F6F9FC}
\newcommand{\hl}{\rowcolor{oursbg}}
\definecolor{cegray}{HTML}{9AA7B6}
\definecolor{scblue}{HTML}{6F9CC4}
\definecolor{stblue}{HTML}{2E5E9E}
\definecolor{uncoral}{HTML}{F09A59}
\definecolor{repgreen}{HTML}{6BAA75}
\definecolor{harmred}{HTML}{E9756F}
\definecolor{chgfill}{HTML}{DCE6F2}
\definecolor{cebandc}{HTML}{E4E8EC}
\newcommand{\grp}[1]{{\color{accent}\bfseries #1}} %
\newcommand{\gpill}[1]{\grp{#1}}
\newcommand{\bandlabel}[1]{{\color{accent}\bfseries #1}} %
\newcommand{\gcmid}[1]{\arrayrulecolor{accent}\cmidrule(lr){#1}\arrayrulecolor{navy}}
\newcommand{\gcmidd}[2]{\arrayrulecolor{accent}\cmidrule(lr){#1}\cmidrule(lr){#2}\arrayrulecolor{navy}}
\newcommand{\gcmiddd}[3]{\arrayrulecolor{accent}\cmidrule(lr){#1}\cmidrule(lr){#2}\cmidrule(lr){#3}\arrayrulecolor{navy}}
\newcommand{\bandrule}{\arrayrulecolor{accentrule}\specialrule{0.4pt}{1.6pt}{1.6pt}\arrayrulecolor{navy}}
\newcommand{\band}[2]{\bandrule\multicolumn{#1}{c}{\bandlabel{#2}}\\\bandrule} %
\newcommand{\firstband}[2]{\multicolumn{#1}{c}{\bandlabel{#2}}\\\bandrule} %
\newcommand{\vsep}{\color{accentrule}\vrule width 0.4pt}
\newcommand{\subrule}{\arrayrulecolor{accentrule!55}\specialrule{0.3pt}{0pt}{0pt}\arrayrulecolor{navy}} %

\newcommand{\tablefont}{\fontsize{8}{9.6}\selectfont} %

\newcommand{\hd}[1]{\multicolumn{1}{l}{#1}} %
\newcolumntype{L}{l} %
\newcolumntype{P}{c}
\newcommand{\oursname}{\textbf{\textsc{ScopeMem}}}
\DeclareRobustCommand{\yes}{\textcolor{repgreen!90!black}{\ding{51}}}
\DeclareRobustCommand{\no}{\textcolor{black!22}{\ding{55}}}
\DeclareRobustCommand{\half}{\tikz[baseline=-0.55ex]{\fill[goldtext] (0,-0.72ex) arc (-90:90:0.72ex) -- cycle; \draw[goldtext,line width=.45pt] (0,0) circle (0.72ex);}}

\DeclareRobustCommand{\stepref}[1]{\tikz[baseline=(c.base)]\node[circle,draw=sky,fill=skypanel,line width=0.45pt,text=navy,inner sep=0.2pt,minimum size=1.0em,font=\tiny\sffamily\bfseries](c){#1};}
\newcommand{\apptab}[1]{Appendix Table~\ref{#1}}

\newcommand{\oursrow}{\textbf{\textsc{ScopeMem} (ours)}}

\title{ScopeMem: Latent Memory That Learns Where to Revise\\Frozen GUI Agents}
\author{%
  Zhicheng Zheng$^{1,*}$ \quad Cheng Zhang$^{1,*}$ \quad
  Shengping Yu$^{2}$ \quad Tianxiang Zhao$^{1}$ \\
  \normalsize $^{1}$The Hong Kong University of Science and Technology (Guangzhou) \\
  \normalsize $^{2}$Imperial College London \\
  \small\texttt{zzheng186@connect.hkust-gz.edu.cn} \\
  \small\texttt{tianxiangz@hkust-gz.edu.cn} \\
  \small $^{*}$Equal contribution.
}

\definecolor{citeblue}{HTML}{2F6EBA}
\hypersetup{citecolor=citeblue, linkcolor=citeblue, urlcolor=citeblue}
\hypersetup{pdftitle={ScopeMem: Latent Memory That Learns Where to Revise Frozen GUI Agents}, pdfauthor={Zhicheng Zheng, Cheng Zhang, Shengping Yu, Tianxiang Zhao}}
\begin{document}
\arrayrulecolor{navy}
\let\oldmidrule\midrule
\renewcommand{\midrule}{\arrayrulecolor{accentrule}\oldmidrule\arrayrulecolor{navy}}
\maketitle
\exhyphenpenalty=10000
\raggedbottom

\begin{abstract}
Latent memory is gaining ground as a lightweight history supplement for GUI agents. It supplies history to a frozen vision-language model via prefix vectors trained with step-wise cross-entropy (CE). While resulting gains are naturally attributed to encoded history, our controlled experiments show they primarily stem from action revision. The prefix learns to alter the model's default actions. However, this revision is unscoped because step-wise CE pushes outputs toward reference actions at every step without explicitly preserving correct decisions. Consequently, it repairs errors but also breaks 8--13\% of initially correct answers. Overall accuracy metrics hide these losses. We therefore propose \textsc{ScopeMem}, a latent-memory method for scoped revision. Its encoder conditions prefix generation on the model's initial answer and is trained to correct errors while preserving correct decisions. On the controlled ablation set, \textsc{ScopeMem} reduces the share of broken correct decisions to 5.6\% while repairing more errors than any evaluated CE-trained prefix. On AndroidControl, GUI-Odyssey, and zero-shot AITW, \textsc{ScopeMem} outperforms all evaluated prompting and latent-memory baselines, enabling a frozen Qwen3-VL-8B to match or surpass GUI-specialized agents and Qwen3-VL-32B.{\let\thefootnote\relax\footnotetext{Code available at \href{https://anonymous.4open.science/r/ScopeMem-1120}{ScopeMem Open-Source Code}}}
\end{abstract}

% Keep paragraph heads and tails together at column breaks.
\begingroup
\clubpenalty=10000
\widowpenalty=10000
\displaywidowpenalty=10000
\section{Introduction}

\begin{figure}[t]
\centering
\includegraphics[width=\columnwidth,trim=14 11 15 7,clip]{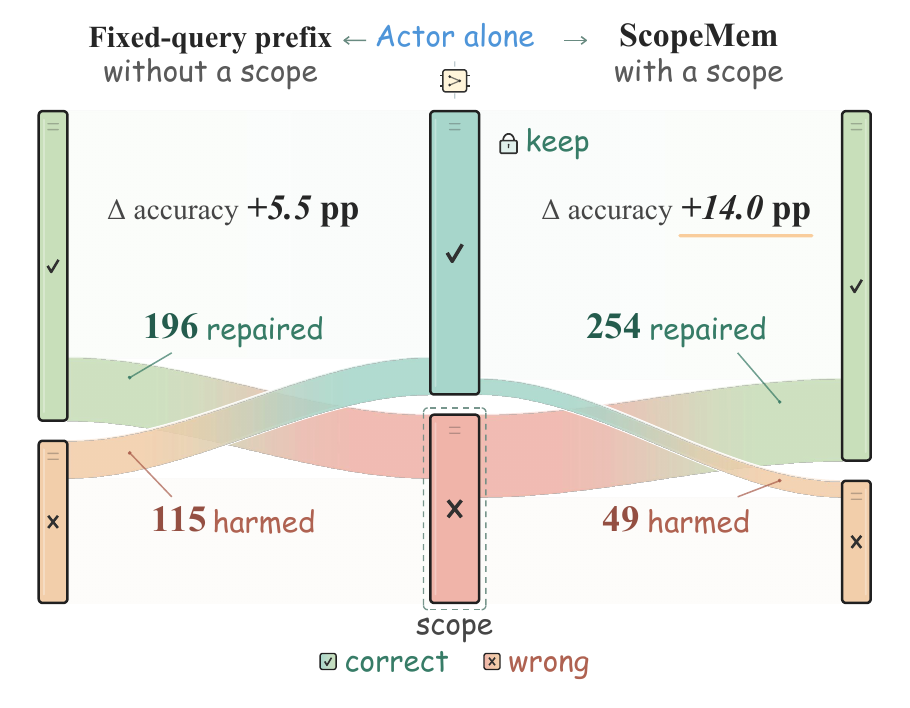}
\caption{Revision cuts both ways. Of the frozen actor's 1,463 decisions (centre), a fixed-query latent-memory prefix (left) repairs 196 errors but harms 115 correct decisions. Its 5.5-point accuracy gain (pp: percentage points) hides these harms. \textsc{ScopeMem} (right) learns to focus revisions on the actor's errors, repairing 254 and harming 49.}
\label{fig:butterfly}
\end{figure}

Latent memory is gaining ground as a lightweight history supplement for GUI agents built on general-purpose vision--language models \citep{wu2025comem,zhang2026memw,zhang2026focusmem}. We call the frozen model that selects each action the \emph{actor}. An encoder converts history into a small set of continuous \emph{prefix} vectors, which are prepended to the actor's input \citep{li2021prefix}. Only the encoder is trained, with step-wise cross-entropy (CE) on reference actions. The prefix's gains are often attributed to stored history. We test this explanation by examining which decisions the prefix changes.

Our controlled experiments (\S\ref{sec:revision}) show that much of the gain persists even when the encoder does not read history. Holding the actor, its context and training fixed, a task-only encoder improves accuracy by 12.0 percentage points, compared with 11.8 points for an encoder that also reads recent history. The actor already sees the same recent history in both conditions, so this comparison isolates the benefit of encoding that history again into the prefix. The small difference suggests that, in this setting, the prefix mainly revises actions using the actor's existing context.

At every step, CE rewards changes that move the actor's output towards the reference action, without requiring its output distribution to remain unchanged when its decision is already correct. This leaves the revision \emph{unscoped}. A change can repair an incorrect decision, but it can also turn a correct decision into an incorrect one, which we call a \emph{harm}. Across our CE-trained prefixes, 8--13\% of the actor's originally correct decisions become incorrect. Encoders that read different inputs exhibit similar harm rates, pointing to their shared objective. Figure~\ref{fig:butterfly} illustrates how overall accuracy gains can conceal many harms because accuracy reports only the net effect of repairs and harms.

We therefore propose \textsc{ScopeMem}, a latent-memory method for \emph{scoped revision}. It aims to repair the actor's errors while preserving the decisions it already gets right. To make this judgement, the encoder needs to see the particular answer it may revise, which a context-only encoder never directly observes. \textsc{ScopeMem} therefore conditions prefix generation on the actor's \emph{first opinion}. Figure~\ref{fig:overview} shows the resulting two-pass process. The frozen actor first responds without a prefix. The encoder then reads this answer and its confidence signals together with the task and recent history, and generates a prefix for a second pass over the same context. The resulting \emph{second opinion} is the action \textsc{ScopeMem} takes.

However, seeing the first opinion does not teach the encoder which decisions to preserve. We therefore train only the encoder with a scoped preservation objective. We call the steps where the actor's original answer is wrong the scope of revision. During training, reference actions identify these steps. CE remains active at every step to encourage the reference action. Outside the scope, we additionally penalise changes to the actor's original output distribution. At deployment, the encoder uses the first opinion and context to decide which answers to preserve, without access to reference actions.

On the controlled ablation set, scoped preservation reduces harms for every encoder, most strongly for the first-opinion encoder. \textsc{ScopeMem} lowers the harm rate to 5.6\% while repairing more errors than any CE-trained prefix. Across AndroidControl, GUI-Odyssey and zero-shot AITW, it outperforms all evaluated prompting and latent-memory baselines. With \textsc{ScopeMem}, the frozen Qwen3-VL-8B actor matches or surpasses GUI-specialised agents and Qwen3-VL-32B under the same actor context.

Our contributions are threefold:
\begin{itemize}[leftmargin=\parindent,topsep=2pt,itemsep=1pt,parsep=0pt]
{\interlinepenalty=10000
\item With recent history already in context, our CE-trained prefixes mainly revise decisions rather than supply history (\S\ref{sec:revision}).\par
}
{\interlinepenalty=10000
\item Cross-entropy leaves this revision unscoped. Every tested prefix trained with CE alone breaks some correct decisions, and aggregate accuracy hides these losses. We therefore recommend reporting repairs and harms alongside accuracy.\par
}
{\interlinepenalty=10000
\item We propose \textsc{ScopeMem}, which combines an encoder that reads the actor's first opinion with a scoped preservation objective. \textsc{ScopeMem} outperforms the evaluated prompting and latent-memory baselines across three benchmarks (\S\S\ref{sec:method}--\ref{sec:analysis}).\par
}
\end{itemize}

\begin{figure*}[t]
\centering
\includegraphics[width=\textwidth]{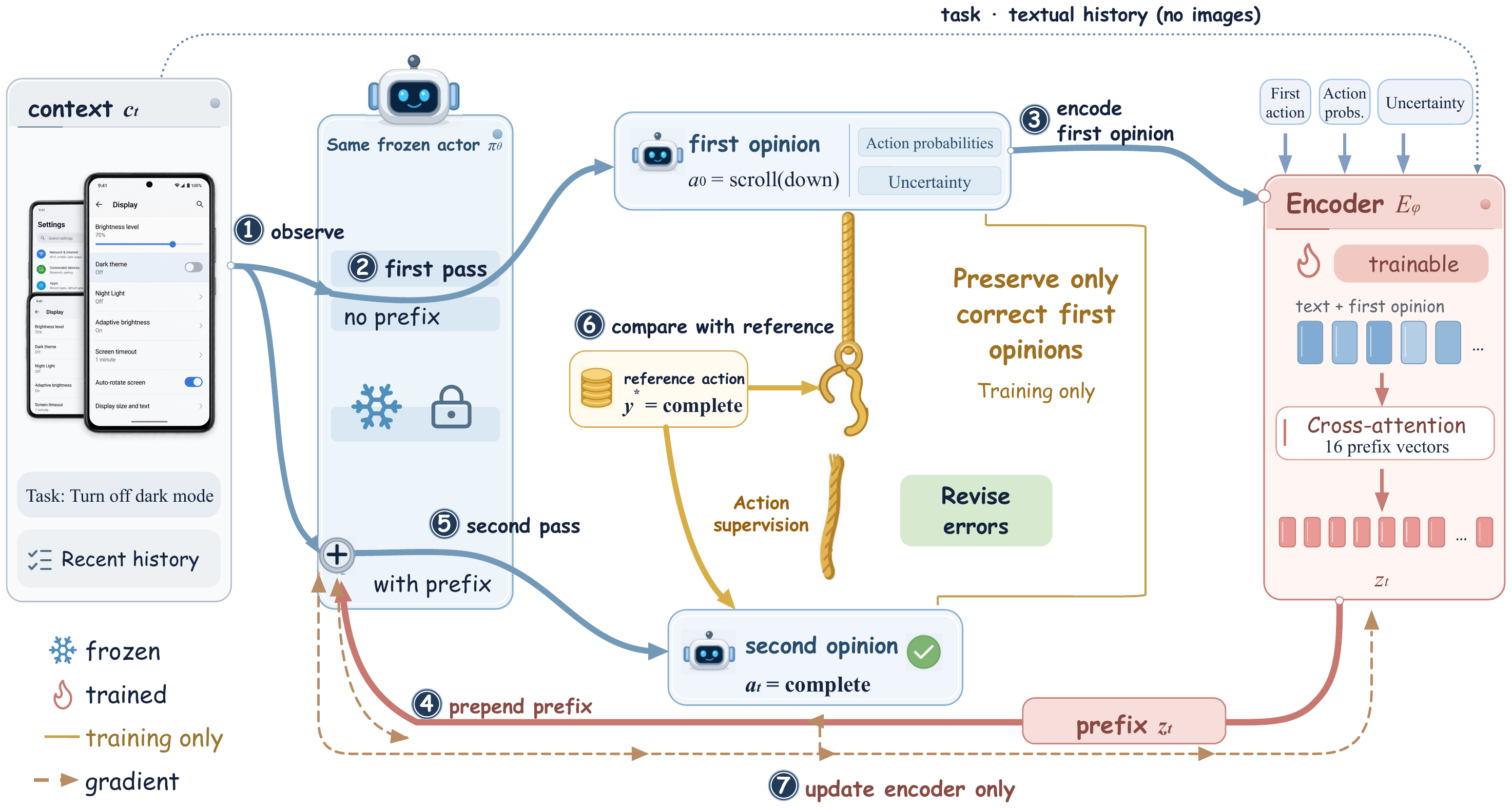}
\caption{\textsc{ScopeMem} gives the frozen actor a second opinion on its first. The actor \stepref{1}~observes the context and \stepref{2}~answers alone in a first pass. The encoder $E_\phi$ \stepref{3}~reads the first opinion, its confidence, the task and textual history to generate a prefix. This prefix is \stepref{4}~prepended to the same context, and the actor \stepref{5}~answers again. Gold parts exist only in training: \stepref{6}~the preservation term is active when the first opinion matches the reference action, and \stepref{7}~gradients update $E_\phi$ only.}
\label{fig:overview}
\end{figure*}

\section{Related Work}

\begin{table}[t]
\centering
\fontsize{8}{9.6}\selectfont
\setlength{\tabcolsep}{3pt}
\resizebox{\columnwidth}{!}{%
\begin{tabular}{l!{\vsep}c!{\vsep}cc!{\vsep}cc}
\toprule
\multirow{2}{*}[-0.4ex]{\textbf{Line of work}} & \gpill{Actor} & \multicolumn{2}{c!{\vsep}}{\gpill{Revision}} & \multicolumn{2}{c}{\gpill{Harms}}\\
\gcmiddd{2-2}{3-4}{5-6}
 & frozen & reads & learned & prevented & reported\\
\midrule
GUI fine-tuning and RL & \no & \no & \yes & \no & \no\\
Explicit memory & \yes & \no & \no & \no & \no\\
Latent memory & \yes & \no & \yes & \no & \no\\
Self-reflection & \yes & \yes & \no & \no & \half\\
\hl \oursname & \yes & \yes & \yes & \yes & \yes\\
\bottomrule
\end{tabular}}
\caption{Only \textsc{ScopeMem} combines learned revision, access to the first opinion, and an objective for preserving correct decisions. \emph{Reads}: sees the first opinion; \emph{learned}: trained; \emph{prevented}: preservation is a training objective; \emph{reported}: repairs and harms reported; \half: partially.}
\label{tab:position}
\end{table}

\paragraph{Vision--language GUI agents.} Vision--language GUI agents learn screenshot-based control through supervised fine-tuning \citep{cheng2024seeclick,wu2025osatlas,xu2024aguvis,qin2025uitars} or reinforcement learning \citep{xu2026mobileagent,yang2026guilibra,venus2026}. To complete multi-step tasks, they must also retain interaction history. Resamplers compress this history \citep{lu2024guiodyssey}, memory units track it \citep{wang2024mobileagentv2}, and summaries retain recent steps \citep{wu2025resum}. Our actor already receives recent history, so we keep that context fixed when evaluating an added prefix.

\paragraph{Memory mechanisms.} Memory supplies reusable information \citep{zhang2026memcut} in two main forms. Explicit memory stores app guidance \citep{zhang2025appagent,wang2025mobileagente}, experiences and workflows \citep{zheng2024synapse,agashe2025agents,wang2025awm}, or task state \citep{zheng2026taskstate}, including hybrid structures \citep{zhu2026helamem,zhu2026hymem}. Latent memory encodes trajectories \citep{wu2025comem,zhang2026memw} or trusted evidence \citep{zhang2026focusmem} into continuous prefixes for a frozen actor \citep{li2021prefix}. LoRA-controlled memory is a parametric alternative \citep{zeng2026memento}. These designs concern how information is represented. We focus on how a latent prefix revises actions, with training that also aims to preserve correct decisions. Table~\ref{tab:position} summarises these distinctions.

\paragraph{Revision and negative flips.} Revision can repair errors but also cause negative flips from correct to incorrect answers. Prompted self-correction revises without training \citep{madaan2023selfrefine} and can introduce as many errors as it repairs without external feedback \citep{huang2024selfcorrect,kamoi2024selfcorrect}. Learned correctors train a model to revise generated answers \citep{welleck2023selfcorrect}. Focal distillation addresses such errors during model updates by preserving predictions the old model gets right \citep{yan2021pct}. \textsc{ScopeMem} applies this preservation principle during prefix training. Its encoder reads the initial answer to guide revision.

\section{Does Latent Memory Remember or Revise?}
\label{sec:revision}

A prefix may supply information the actor lacks or change how it uses information already in its context. To distinguish these explanations, we \mbox{withhold} history from the encoder while keeping the actor's history available. If the prefix helps mainly by encoding history, its gain should shrink. If it mainly revises how the actor uses existing context, much of the gain should remain.

Revision also predicts a side effect. Since cross-entropy pulls every step towards the reference action whatever the prefix reads, a revising prefix should change correct decisions as well as wrong ones. Prefixes trained in the usual way, with cross-entropy alone, behave as revision predicts on both counts.

\subsection{A controlled study}
\label{sec:formulation}

At step $t$ the frozen actor $\pi_\theta$ reads a \emph{history-memory context} $c_t$ and emits the next action. This context consists of the task, a rolling summary of the last three steps and the latest three screenshots. A \emph{prefix} $z_t$ of $K{=}16$ vectors is prepended to the actor's input embeddings. Only the module that writes the prefix is trained.

\noindent The actor then selects an action as
\begingroup\postdisplaypenalty=10000
\begin{equation}
a_t \sim \pi_\theta(\cdot\mid z_t\oplus c_t),
\label{eq:prefix}
\end{equation}
\endgroup
where $\oplus$ denotes prepending. We compare three prefixes with different inputs. The \emph{fixed-query prefix}, used in latent memory \citep{zhang2026memw,zhang2026focusmem}, uses one set of learned queries shared across steps to read the current context inside the actor's forward pass. The \emph{task-only encoder} reads the task, while the \emph{history encoder} also reads the rolling summary and recent actions. All three use the same actor, actor context, training data and schedule. They minimise cross-entropy on the \emph{reference action} $y^*$ annotated for each step. We compare each with the \emph{plain actor}, which has no prefix. Evaluation uses a held-out \emph{ablation set} of 1,463 test steps from AndroidControl (high-level instructions) and GUI-Odyssey (random split). We report paired 95\% trajectory-bootstrap intervals for differences between methods. Appendix~\ref{app:ablation} reports the key paired contrasts.

\begin{figure}[t]
\centering
\includegraphics[width=\columnwidth]{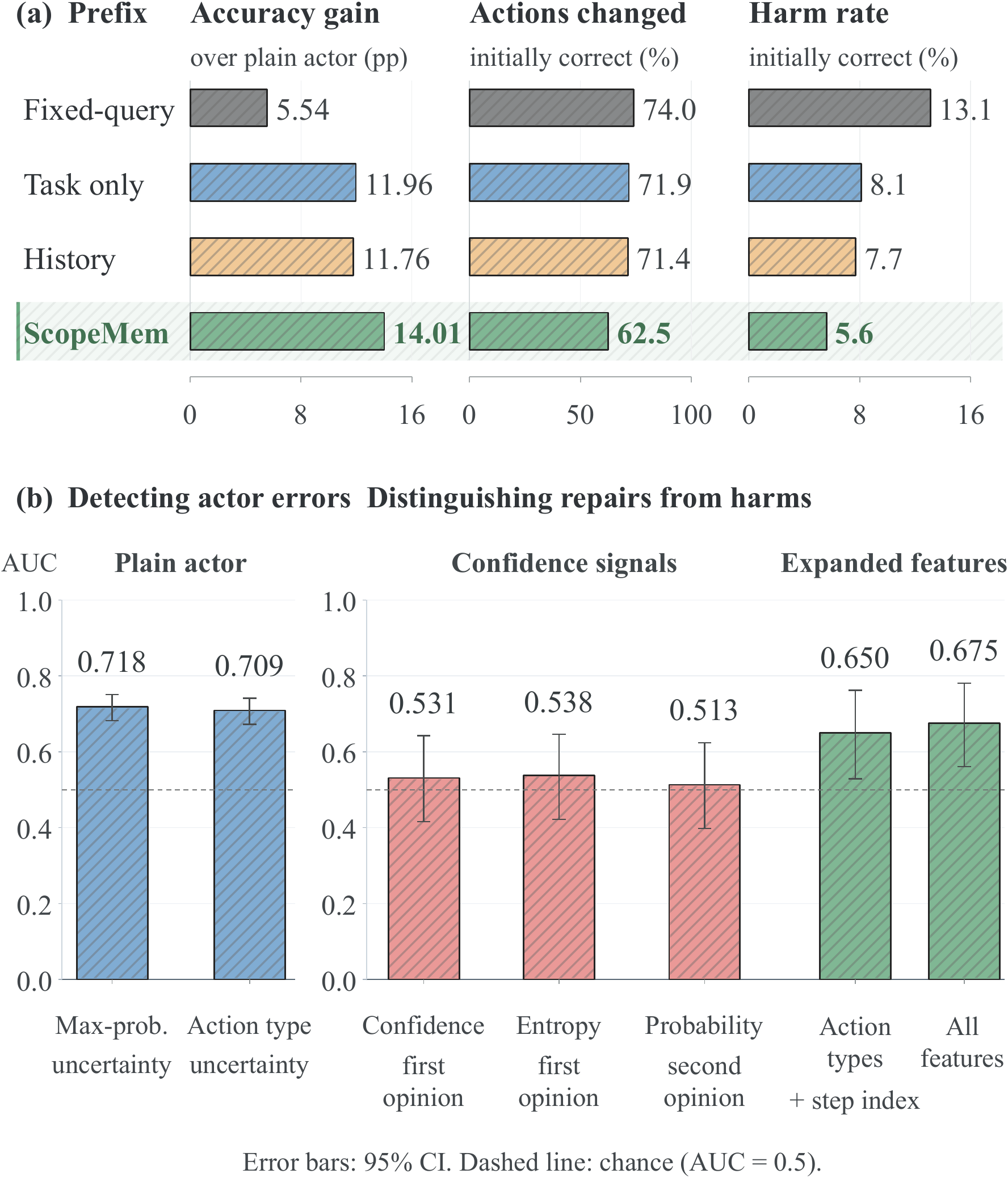}
\caption{Prefixes trained with cross-entropy alone also change correct decisions, and confidence barely separates repairs from harms. (a) Ablation set: gain over the plain actor (pp), share of correct actions changed, and harm rate. Shading marks \textsc{ScopeMem}. (b) AUC of confidence signals and expanded features for detecting the plain actor's errors and separating \textsc{ScopeMem}'s repairs from harms.}
\label{fig:teaser}
\end{figure}

\subsection{The gain comes from revision}
\label{sec:nomemory}

Figure~\ref{fig:teaser}a shows that the task-only encoder improves the actor as much as the history encoder (12.0 vs.\ 11.8\,pp), and both improve it about twice as much as the fixed-query prefix. For history minus task-only accuracy, the paired 95\% confidence interval has an upper bound of 1.6\,pp, a small fraction of either encoder's gain. The prefix thus mainly revises how the actor uses its existing context and should be judged by the decisions it changes.

\subsection{The revision repairs and harms}
\label{sec:harms}

Accuracy measures only the net effect of a revision, so we count its two kinds of change separately. A \emph{repair} is a step that the plain actor gets wrong and the prefixed actor gets right, and a \emph{harm} is the reverse. With R repairs and H harms over $N$ steps, the \emph{gain} over the plain actor is
\begingroup\postdisplaypenalty=10000
\begin{equation}
\Delta\mathrm{Acc}=\frac{\mathrm{R}-\mathrm{H}}{N}.
\label{eq:ledger}
\end{equation}
\endgroup
We also report two rates. The \emph{repair rate} divides R by the plain actor's errors, and the \emph{harm rate} divides H by its correct steps.

Every tested prefix changes most of the plain actor's correct actions in at least one field. Changes need not cause errors, since coordinate shifts within the benchmark's tolerance still count as changes. Most modified actions remain correct, but some become incorrect. These harms cancel more than a quarter of what each encoder repairs, and more than half for the fixed-query prefix. The task-only and history encoders harm at nearly the same rate although they read different inputs, which points to the objective they share.

A gate could fall back to the plain actor's answer to avoid harms, but it would need to distinguish harmful from helpful revisions. The actor's confidence detects its own errors moderately well (AUC 0.72). However, Figure~\ref{fig:teaser}b shows that this confidence barely distinguishes \textsc{ScopeMem}'s repairs from harms (AUC 0.53). This motivates learning the scope rather than gating revisions by this confidence alone.

\begin{table*}[t]
\centering
\footnotesize
\setlength{\tabcolsep}{5.4pt}
\newcommand{\dg}[1]{{\scriptsize\color{repgreen!70!black}#1}}
\newcommand{\on}{\hspace{0.6em}$\hookrightarrow$\hspace{0.3em}}
\begin{tabular}{l!{\vsep}cc!{\vsep}cc!{\vsep}ccccc@{\hspace{9pt}}c}
\toprule
\multirow{2}{*}[-0.4ex]{\textbf{Method}} & \multicolumn{2}{c!{\vsep}}{\grp{AndroidControl}} & \multicolumn{2}{c!{\vsep}}{\grp{GUI-Odyssey}} & \multicolumn{6}{c}{\grp{AITW (zero-shot)}}\\
\gcmiddd{2-3}{4-5}{6-11}
 & EM & TM & EM & TM & Gen. & Inst. & Google & Single & Shop & Macro\\
\midrule
\firstband{11}{General-purpose models}
Qwen3-VL-4B & 60.64 & 76.42 & 57.18 & 78.06 & 54.19 & 68.05 & 57.40 & 66.67 & 53.44 & 59.95\\
Qwen3-VL-32B & 61.88 & 78.37 & 61.84 & 80.30 & 57.54 & 69.82 & 62.13 & 63.16 & 56.87 & 61.90\\
InternVL3.5-8B & 35.28 & 57.80 & 29.52 & 60.41 & 40.22 & 39.64 & 28.99 & 47.37 & 34.35 & 38.12\\
GLM-4.1V-9B & 40.96 & 70.74 & 35.80 & 67.56 & 37.43 & 44.97 & 39.05 & 40.35 & 35.88 & 39.54\\
\band{11}{Trained for GUI control}
Aguvis-7B & 64.18 & 81.91 & 40.96 & 64.08 & 41.90 & 40.24 & 32.54 & 33.33 & 41.60 & 37.92\\
UI-TARS-1.5-7B & 50.00 & 72.70 & 37.10 & 62.15 & 45.25 & 65.68 & 46.75 & 54.39 & 39.31 & 50.28\\
UI-Venus-1.5-8B & 63.48 & 75.18 & 60.97 & 76.69 & 60.89 & 68.64 & 63.31 & 59.65 & 58.40 & 62.18\\
GUI-Owl-1.5-8B & 67.55 & 82.62 & 60.66 & 76.07 & 52.51 & 55.03 & 56.80 & 50.88 & 51.15 & 53.27\\
GUI-Libra-8B & 65.43 & 79.08 & 57.18 & 73.90 & 49.16 & 65.09 & 50.30 & 71.93 & 50.38 & 57.37\\
\band{11}{Prompting or memory on the frozen actor}
Qwen3-VL-8B (plain actor) & 62.59 & 75.35 & 58.73 & 78.74 & 60.89 & 65.68 & 57.99 & 54.39 & 52.29 & 58.25\\
\on Self-Reflection & 57.98 & 68.97 & 47.98 & 67.31 & 58.66 & 67.46 & 59.76 & 57.89 & 47.71 & 58.30\\
\on ReAct & 63.30 & 76.42 & 62.34 & 79.80 & 62.01 & 67.46 & 59.17 & 59.65 & 53.05 & 60.27\\
\on ReSum & 63.83 & 76.95 & 63.70 & 80.36 & \textbf{62.57} & 68.64 & 59.76 & 63.16 & 53.82 & 61.59\\
\subrule
\on CoMEM$^\dagger$ & 63.30 & 78.19 & 62.77 & 81.48 & 58.66 & 68.64 & 59.76 & 68.42 & 56.11 & 62.32\\
\on HyMEM & 64.89 & 78.37 & 65.26 & 82.41 & 62.01 & 68.05 & 60.95 & 63.16 & 56.11 & 62.05\\
\on Mem-W & 65.96 & 80.32 & 68.61 & 84.03 & 62.01 & 69.23 & 62.72 & 68.42 & 57.63 & 64.00\\
\hl \on \oursrow & \textbf{67.91} & \textbf{82.98} & \textbf{72.90} & \textbf{86.33} & \textbf{62.57} & \textbf{71.01} & \textbf{64.50} & \textbf{75.44} & \textbf{59.92} & \textbf{66.69}\\
\hl \dg{\hspace{2.1em}gain over the plain actor} & \dg{+5.32} & \dg{+7.63} & \dg{+14.17} & \dg{+7.59} & \dg{+1.68} & \dg{+5.33} & \dg{+6.51} & \dg{+21.05} & \dg{+7.63} & \dg{+8.44}\\
\bottomrule
\end{tabular}
\caption{On the frozen actor, only \textsc{ScopeMem} matches or beats GUI-specialised models on all three benchmarks. Step accuracy (\%) on the main test sets; EM: full action; TM: action type; AITW subsets General, Install, GoogleApps, Single, WebShopping and their mean (Macro). $\hookrightarrow$: on the frozen plain actor ($^\dagger$: on Qwen2.5-VL-7B); last row: gain in pp.}
\label{tab:main}
\end{table*}

\section{\textsc{ScopeMem}: Latent Memory as a Scoped Revision}
\label{sec:method}

Our ideal revision repairs every error and preserves every correct decision, maximising Eq.~\eqref{eq:ledger}. Let $a_0$ be the actor's answer without a prefix and $\pi_0$ the output distribution used to decode that answer.

\noindent The ideal distribution with the prefix is
\begingroup\postdisplaypenalty=10000
\begin{equation}
\pi_z^\star=
\begin{cases}
\delta_{y^*} & \text{if } a_0 \text{ is wrong},\\
\pi_0 & \text{if } a_0 \text{ is right},
\end{cases}
\label{eq:ideal}
\end{equation}
\endgroup
where $\delta_{y^*}$ puts all mass on the reference action. The \emph{scope} consists of steps on which $a_0$ is wrong. The ideal revision corrects these errors and leaves the original output distribution unchanged on correct steps. Keeping this distribution unchanged also preserves the correct answer. \textsc{ScopeMem} learns to approximate this ideal. Its encoder reads the first opinion to guide revision at deployment, and its training objective encourages correction inside the scope and preservation outside it.

\subsection{A second opinion on the actor's answer}
\label{sec:encoder}

Under Eq.~\eqref{eq:ideal}, revision is needed when the actor's initial answer is wrong. An encoder without screenshots can use textual context to estimate the actor's chance of being correct. However, that context does not show the answer the actor actually produced. We therefore first query the frozen actor without a prefix to obtain the \emph{first opinion} $a_0$. The encoder receives this answer before the second actor pass. Both passes use greedy decoding.

The \emph{confidence inputs} are the actor's probabilities over ten action types ($p_0$) and their entropy ($H_0$). The \textbf{first-opinion encoder} $E_\phi$ is a 10.5M-parameter cross-attention network. It generates $z_t$ from $a_0$, these inputs, the task, rolling summary and recent actions. With this prefix, the frozen actor produces the \emph{second opinion} $a_t$, \textsc{ScopeMem}'s selected action.

\subsection{Learning the scope}
\label{sec:scoped}

Reference actions are unavailable at deployment, so correctness labels guide the encoder only during training. The \emph{keep indicator} $k_t$ is 1 when the initial answer is correct under the benchmark's matching rules, and 0 when the answer is wrong. The training objective is
\begingroup\postdisplaypenalty=10000
\begin{equation}
\mathcal{L} = \mathrm{CE} + k_t\,w_t\,\mathcal{P}_t,
\label{eq:loss}
\end{equation}
\endgroup
where $\mathrm{CE}$ is the cross-entropy loss of the reference action and $\mathcal{P}_t$ is a \emph{preservation term} based on KL divergence. For incorrect initial answers ($k_t=0$), CE encourages the reference action. For correct initial answers, weighted preservation penalises differences between the actor's token distributions with and without the prefix, evaluated along the reference action. CE still encourages the reference tokens, while preservation also constrains changes to other token probabilities. We call this objective \textbf{scoped preservation}. The encoder receives no correctness label $k_t$. Instead, the encoder learns to generate prefixes from the first opinion, context and confidence inputs.

Three choices shape the objective. First, benchmarks require exact action-type matches but allow tolerance for arguments such as coordinates. We therefore give \emph{decision tokens}, including those specifying action type, ten times the preservation weight of \emph{argument tokens}. Second, the confidence weight $w_t$ increases with confidence, so confident correct answers receive stronger preservation. Third, the first epoch uses $\mathrm{CE}$ alone to teach the prefix to revise. Adding preservation immediately would reward a prefix that changes nothing. The second epoch adds preservation to teach the encoder which decisions to keep.

\section{Experiments}
\label{sec:experiments}

\subsection{Setup}
\label{sec:setup}

\paragraph{Benchmarks.} We use three standard offline benchmarks for mobile GUI control. \textbf{AndroidControl} \citep{li2024androidcontrol} covers everyday tasks within an app, and \textbf{\mbox{GUI-Odyssey}} \citep{lu2024guiodyssey} covers long cross-app tasks, where history should matter most. Prefixes are trained on the training splits of these two benchmarks. \textbf{AITW} \citep[Android in the Wild;][]{rawles2023aitw} is held out from training and tests transfer. We sample whole trajectories from the official test splits to form the \textbf{main test sets}: 564 steps for AndroidControl, 1,609 for GUI-Odyssey and 836 for AITW. Standard scorers report full-action (EM) and action-type (TM) accuracy for AndroidControl and GUI-Odyssey, and subset scores with their macro average for AITW.

\paragraph{Baselines.} An actor can be improved in four ways, and we compare with recent methods of each. To \emph{scale} or replace the general-purpose model, we use Qwen3-VL at 4B and 32B around our 8B actor, InternVL3.5-8B and GLM-4.1V-9B \citep{bai2025qwen3vl,wang2025internvl,glm2025}. Aguvis, UI-TARS-1.5, UI-Venus-1.5, GUI-Owl-1.5 and GUI-Libra represent \emph{training} for GUI control \citep{xu2024aguvis,qin2025uitars,venus2026,xu2026mobileagent,yang2026guilibra}. The \emph{prompting} baselines are ReAct \citep{yao2023react}, ReSum \citep{wu2025resum} and Self-Reflection, which revises the answer verbally in the manner of Self-Refine \citep{madaan2023selfrefine}. Among methods that add \emph{memory}, we compare with CoMEM \citep{wu2025comem}, HyMEM \citep{zhu2026hymem} and Mem-W \citep{zhang2026memw}, which is the closest to \textsc{ScopeMem}. The memory baselines retain their method-specific \mbox{memory} inputs and use \textsc{ScopeMem}'s training data and schedule. All non-memory baselines receive the same actor context as \textsc{ScopeMem}.

\begin{table}[t]
\centering
\tablefont
\setlength{\tabcolsep}{2.2pt}
\resizebox{\columnwidth}{!}{%
\begin{tabular}{lcccc@{\hspace{8pt}}c}
\toprule
\multirow{2}{*}[-0.4ex]{\textbf{Benchmark}} & \multicolumn{2}{c}{\gpill{Repairs / harms}} & \multicolumn{2}{c}{\gpill{Only one system}} & \multirow{2}{*}[-0.4ex]{Advantage (95\% CI)}\\
\gcmidd{2-3}{4-5}
 & \textsc{ScopeMem} & Mem-W & keeps & repairs & \\
\midrule
AndroidControl & 57/27 & 55/36 & 16/7 & 13/11 & +1.95 [$-$0.56, 4.62]\\
GUI-Odyssey & 264/36 & 218/59 & 33/10 & 71/25 & +4.29 [2.76, 5.84]\\
AITW & 87/32 & 82/44 & 17/5 & 15/10 & +2.03 [0.36, 3.77]\\
\bottomrule
\end{tabular}}
\caption{Mem-W breaks more correct decisions than \textsc{ScopeMem}. Counts on the main test sets are relative to the plain actor. ``Only one system'' counts correct steps only one preserves or errors only one repairs (\textsc{ScopeMem}/Mem-W). Advantage: \textsc{ScopeMem}'s accuracy minus Mem-W's (pp, 95\% CI).}
\label{tab:memw}
\end{table}

\begin{table}[t]
\centering
\tablefont
\setlength{\tabcolsep}{2.6pt}
\begin{tabular}{l!{\vsep}rr!{\vsep}rr!{\vsep}rr}
\toprule
\multirow{2}{*}[-0.4ex]{\textbf{Second epoch}} & \multirow{2}{*}[-0.4ex]{R} & \multirow{2}{*}[-0.4ex]{H} & \multicolumn{2}{c!{\vsep}}{\gpill{Rate (\%)}} & \multirow{2}{*}[-0.4ex]{Acc} & \multirow{2}{*}[-0.4ex]{Gain}\\
\gcmid{4-5}
 & & & repair & harm & & \\
\midrule
\firstband{7}{Fixed-query prefix}
cross-entropy alone & 196 & 115 & 33.6 & 13.1 & 65.62 & +5.54\\
\band{7}{Task-only encoder}
cross-entropy alone & 246 & 71 & 42.1 & 8.1 & 72.04 & +11.96\\
scoped & 242 & 58 & 41.4 & 6.6 & 72.66 & +12.58\\
\band{7}{History encoder}
first epoch only & 245 & 74 & 42.0 & 8.4 & 71.77 & +11.69\\
cross-entropy alone & 240 & 68 & 41.1 & 7.7 & 71.84 & +11.76\\
scoped & 235 & 53 & 40.2 & 6.0 & 72.52 & +12.44\\
\band{7}{First-opinion encoder}
first epoch only & 244 & 70 & 41.8 & 8.0 & 71.98 & +11.89\\
cross-entropy alone & 243 & 67 & 41.6 & 7.6 & 72.11 & +12.03\\
unscoped & 212 & 47 & 36.3 & 5.3 & 71.36 & +11.28\\
\hl \textbf{scoped (\textsc{ScopeMem})} & 254 & 49 & 43.5 & 5.6 & 74.09 & +14.01\\
\bottomrule
\end{tabular}
\caption{Scoped preservation lowers harms for every encoder. Matched variants on the ablation set (\emph{first epoch only}: checkpoint before the second epoch). Rates: R over 584 errors, H over 879 correct steps; Acc: accuracy (\%); Gain: over the plain actor (pp).}
\label{tab:ablation}
\end{table}

\subsection{A frozen open-source actor matches GUI-specialised models}
\label{sec:main}

Table~\ref{tab:main} shows that, among methods on the frozen actor, only \textsc{ScopeMem} matches the strongest model \emph{trained for GUI control} on each benchmark. On GUI-Odyssey, prompting and memory methods other than Self-Reflection already exceed the strongest GUI-trained baseline in EM. CoMEM and Mem-W also exceed the strongest GUI-trained baseline on AITW Macro. On AndroidControl, only \textsc{ScopeMem} reaches the best GUI-trained model, GUI-Owl (67.91 vs.\ 67.55 EM). On the held-out AITW, \textsc{ScopeMem} ties or leads every method on every subset.

\emph{Scaling} helps less evenly. Relative to the plain 8B actor, the four-fold larger Qwen3-VL-32B gains 3.1\,pp EM on GUI-Odyssey and 3.7\,pp Macro on AITW, but loses 0.7\,pp EM on AndroidControl. \textsc{ScopeMem}'s 10.5M-parameter module gains 5.3--14.2\,pp over the same actor. Against \emph{prompting and memory}, each of the three \textsc{ScopeMem} training runs in Appendix~\ref{app:seeds} exceeds every other method on the frozen actor. Self-Reflection also reads the actor's draft and spends a second actor pass, but its revision is prompted rather than learned, and it lowers accuracy on two of the three benchmarks. Reading the first opinion is therefore insufficient without learning which answers to revise (\S\ref{sec:fo}).

\subsection{Where \textsc{ScopeMem} gains over latent memory}
\label{sec:memw}

Mem-W is the strongest memory method. It writes a prefix for the same frozen actor, using the same data and schedule. The prefix holds retrieved past trajectories and a working snapshot of the current one, and Mem-W trains it with cross-entropy and self-distillation on every step.

\begin{figure}[t]
\centering
\includegraphics[width=\columnwidth,trim=38 15 13 23,clip]{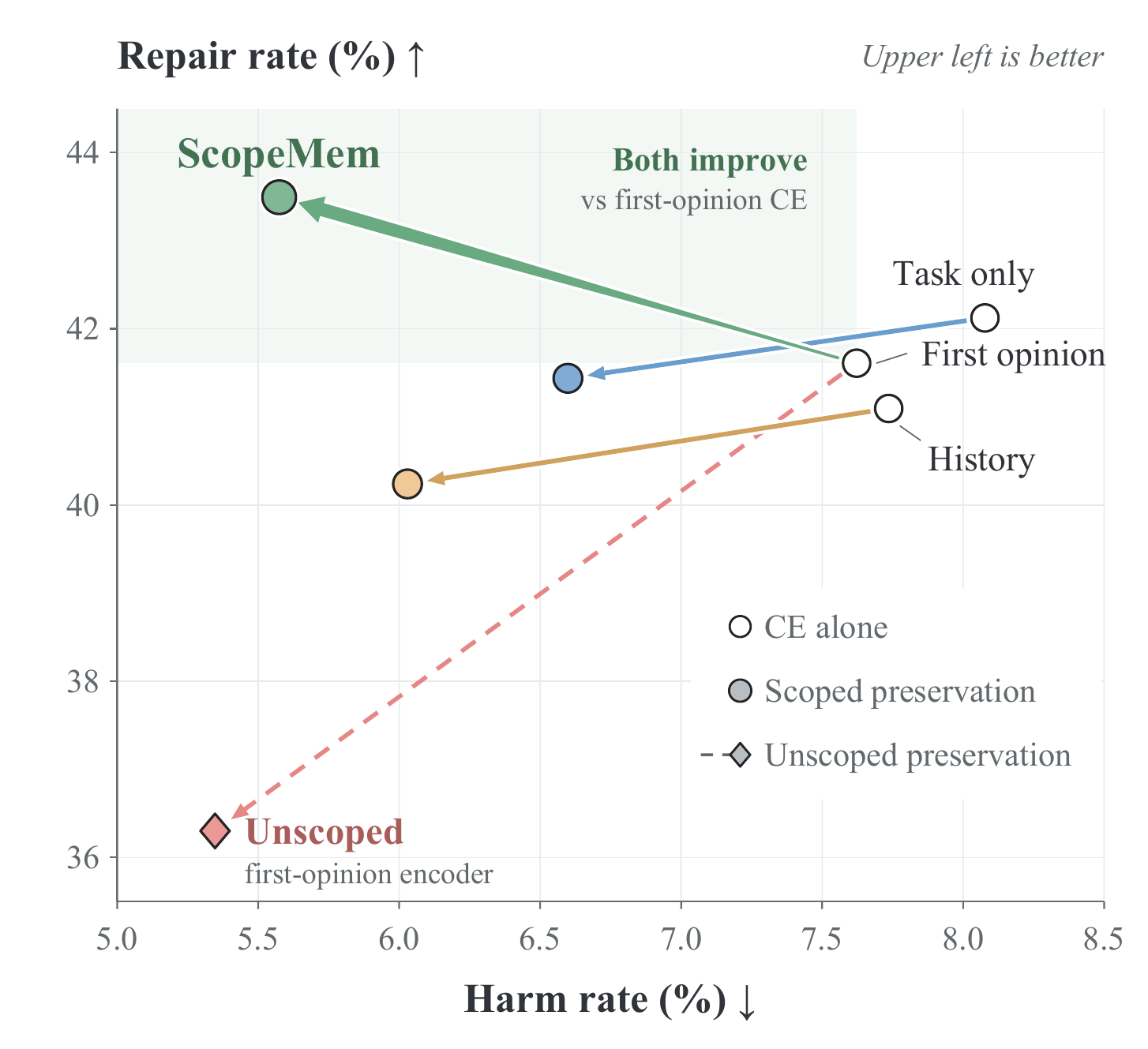}
\caption{On the ablation set, arrows compare the same encoder trained with cross-entropy alone and with scoped (solid) or unscoped (dashed) preservation. Only \textsc{ScopeMem} lowers harms and raises repairs together, reaching the shaded region. The fixed-query prefix (harm rate 13.1\%, repair rate 33.6\%) lies outside the plotted range.}
\label{fig:plane}
\end{figure}

Table~\ref{tab:memw} shows the most consistent difference in harms: Mem-W breaks a third to nearly two thirds more correct decisions than \textsc{ScopeMem}. On AndroidControl and AITW, the methods repair similar numbers of errors, so \textsc{ScopeMem}'s accuracy advantage comes mainly from preserving more correct decisions. The AndroidControl accuracy gap remains within sampling error. The \mbox{advantage} is largest on GUI-Odyssey, where \textsc{ScopeMem} also repairs 46 more errors. GUI-Odyssey has the longest trajectories, where memory of earlier steps could be especially useful. However, \textsc{ScopeMem}'s advantage over Mem-W is similar on short and long trajectories, consistent with revision as its main source. Much of this advantage comes from completion decisions, where our Mem-W reproduction more often ends tasks too early, and from clicks, which account for most errors that only \textsc{ScopeMem} repairs.

\section{Analysis: Testing the Scope}
\label{sec:analysis}

We test three predictions of scoped revision with matched variants. (i)~Preservation outside the scope lowers harms regardless of encoder input. (ii)~Preserving every step also holds the actor to its errors. (iii)~Scoping helps the first-opinion encoder most because it observes the decision being revised. Each band of Table~\ref{tab:ablation} fixes the prefix's inputs and varies only the second epoch. This epoch is skipped (\emph{first epoch only}) or uses \emph{cross-entropy alone}, \emph{scoped preservation} or \emph{unscoped preservation}. The last applies preservation at every step ($k_t=1$).

\subsection{Does preservation remove harms, and at what cost?}
\label{sec:ablation}

Switching the second epoch from cross-entropy alone to scoped preservation removes 17--27\% of each encoder's harms in every training run, as claim~(i) predicts and Figure~\ref{fig:plane} shows for the main run. Part of the harms in \S\ref{sec:harms} therefore comes from the objective, whatever the prefix reads.

Preserving every step costs repairs (claim~ii). For the first-opinion encoder, unscoped preservation reduces harms about as much as scoped preservation. Relative to cross-entropy alone, however, it loses one in eight repairs and lowers accuracy. Preserving outputs at every step also discourages correction of the actor's mistakes, for instance on completion steps. Under scoped preservation, the task-only and history encoders retain almost all their CE repairs. The first-opinion encoder is the only variant that both lowers harms and raises repairs relative to CE alone.

Neither more training nor the second actor pass explains the gain. Starting from the first-epoch checkpoint, a second epoch of cross-entropy alone barely changes the first-opinion encoder's accuracy. Scoped preservation in the second epoch raises accuracy by 2.1\,pp over that same checkpoint, beyond sampling error. With cross-entropy alone, the two-pass first-opinion encoder and the one-pass task-only encoder gain similar amounts over the plain actor (12.03 vs.\ 11.96\,pp), with the difference within sampling error.

Most of the harms the scope removes are action-type harms, including premature completions, as Appendix~\ref{app:harms} shows. The action type is chosen by the decision tokens, which the preservation term weights most.

\subsection{Why does the scope need the first opinion?}
\label{sec:fo}

\apptab{tab:contrasts} shows that scoped preservation adds less than one percentage point of accuracy over CE alone for the task-only and history encoders, but about three times as much for the first-opinion encoder. \apptab{tab:decomp} shows similar reductions in the shares of correct actions and errors changed by the task-only and history encoders under scoped preservation. Reducing revision overall can avoid harms, but it can also prevent useful corrections.

For the first-opinion encoder, scoping reduces changes to correct actions by a similar amount as for the other two encoders. However, the scoped encoder changes as many of the actor's errors as its CE version and repairs more of them. This selectivity supports the motivation in \S\ref{sec:encoder}: the encoder observes the actual answer being revised. Across the three training runs in Appendix~\ref{app:seeds}, the accuracy gain from scoped preservation over CE is larger for the first-opinion encoder than for the history encoder. This interaction averages 1.18\,pp (95\% interval across runs [0.38, 1.99]). For the main run alone, the bootstrap interval for the interaction includes zero.

\paragraph{What the prefix uses.} The interventions on the encoder's inputs in \apptab{tab:prefix} point the same way. Corrupting the first opinion lowers accuracy beyond sampling error, whereas removing history changes it only within sampling error.

\subsection{Case study}
\label{sec:case}

\begin{figure}[t]
\centering
\includegraphics[width=\columnwidth,trim=0 1 0 2,clip]{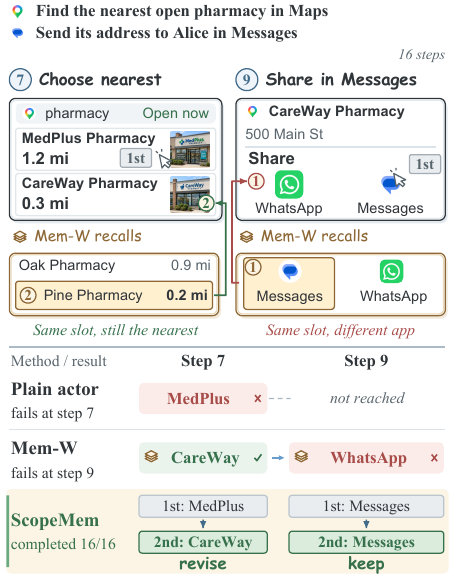}
\caption{Case study: one recalled episode repairs step~7 and breaks step~9 for Mem-W, whereas \textsc{ScopeMem} revises step~7 and keeps step~9.}
\label{fig:case}
\end{figure}

At step~7 of the test task in Figure~\ref{fig:case}, Mem-W selects the second list item, the position of the nearest pharmacy in a recalled episode, and repairs the plain actor's error. At step~9, Mem-W again follows the recalled interface and selects the first share option. That position held Messages in the old episode but holds WhatsApp in the current task, so Mem-W breaks the plain actor's correct \mbox{decision}. The same recalled episode repairs one step and harms another despite appearing equally relevant at both. \textsc{ScopeMem} revises the incorrect first opinion at step~7, preserves the correct one at step~9 and gets all 16 steps right.

\section{Conclusion}

We study latent memory as a history supplement for frozen GUI agents and show that, when recent history is already available, much of its benefit comes from revising decisions. Our controlled experiments reveal that CE-trained prefixes repair errors but also break correct decisions, losses that aggregate accuracy hides. We propose \textsc{ScopeMem} to scope this revision. Its encoder reads the actor's first answer and learns to generate prefixes that revise errors while preserving correct answers. \textsc{ScopeMem} achieves more repairs and fewer harms than CE-trained prefixes in controlled ablations. Across three benchmarks, it outperforms prompting and latent-memory baselines and enables a frozen 8B actor to match or surpass GUI-specialised models.

\clearpage
\endgroup
\section*{Limitations}

Our experiments are offline and step-level, so online task success, for example on AndroidWorld \citep{rawles2025androidworld}, remains to be tested. Step-level matching may also penalise valid alternatives to the reference actions, and noise in these references mislabels some first opinions during training. We study one actor, Qwen3-VL-8B, on mobile benchmarks, and our analysis of memory concerns an actor that already sees its recent steps. Other actors, platforms and history budgets may behave differently. The memory baselines follow their original papers under our shared training budget, and our fine-grained analyses come from one training run. \textsc{ScopeMem} also queries the actor twice per step, which roughly doubles its latency. We leave online evaluation and memories that reach beyond the recent steps to future work.

\section*{Ethical Considerations}

We use existing public GUI benchmarks, without new participant recruitment or online action execution. Screenshots may contain names, account identifiers or other incidental personal information, and we have not completed a de-identification audit. Frozen-actor interventions could in principle be used to steer an agent towards unintended actions. Our method changes only a small module and is evaluated offline.

\paragraph{Assets and intended use.} The data and models serve offline research, and our system is a research prototype. GUI-Odyssey is released under CC BY 4.0; Qwen3-VL-4B/8B and Qwen2.5-VL-7B model cards specify Apache 2.0. AndroidControl and AITW are released by Google Research through a repository under the Apache 2.0 licence, with no separate data licence, and the third-party rights of individual screenshots lie outside it. We provide source links and remaining licence-verification limits for these and the other models in the accompanying Responsible NLP statement, and we assert no new redistribution licence for third-party screenshots or model weights.

\section*{Use of AI Assistants}

We used generative AI assistants only to polish the wording of this paper. The research questions, the design of the method and the experiments, the analyses, and every figure, table, and number were produced by the authors.

\bibliography{custom}

@inproceedings{agashe2025agents,
  author = {Saaket Agashe and Jiuzhou Han and Shuyu Gan and Jiachen Yang and Ang Li and Xin Wang},
  title = {Agent {S}: An open agentic framework that uses computers like a human},
  booktitle = {The Thirteenth International Conference on Learning Representations},
  year = {2025}
}

@article{bai2025qwen3vl,
  author = {Shuai Bai and Yuxuan Cai and Ruizhe Chen and Keqin Chen and Xionghui Chen and Zesen Cheng and Lianghao Deng and Wei Ding and Chang Gao and Chunjiang Ge and Wenbin Ge and Zhifang Guo and Qidong Huang and Jie Huang and Fei Huang and Binyuan Hui and Shutong Jiang and Zhaohai Li and Mingsheng Li and Mei Li and Kaixin Li and Zicheng Lin and Junyang Lin and Xuejing Liu and Jiawei Liu and Chenglong Liu and Yang Liu and Dayiheng Liu and Shixuan Liu and Dunjie Lu and Ruilin Luo and Chenxu Lv and Rui Men and Lingchen Meng and Xuancheng Ren and Xingzhang Ren and Sibo Song and Yuchong Sun and Jun Tang and Jianhong Tu and Jianqiang Wan and Peng Wang and Pengfei Wang and Qiuyue Wang and Yuxuan Wang and Tianbao Xie and Yiheng Xu and Haiyang Xu and Jin Xu and Zhibo Yang and Mingkun Yang and Jianxin Yang and An Yang and Bowen Yu and Fei Zhang and Hang Zhang and Xi Zhang and Bo Zheng and Humen Zhong and Jingren Zhou and Fan Zhou and Jing Zhou and Yuanzhi Zhu and Ke Zhu},
  title = {{Qwen3-VL} technical report},
  journal = {arXiv preprint arXiv:2511.21631},
  year = {2025},
  doi = {10.48550/arXiv.2511.21631}
}

@inproceedings{cheng2024seeclick,
  author = {Kanzhi Cheng and Qiushi Sun and Yougang Chu and Fangzhi Xu and Li YanTao and Jianbing Zhang and Zhiyong Wu},
  title = {{SeeClick}: Harnessing {GUI} grounding for advanced visual {GUI} agents},
  booktitle = {Proceedings of the 62nd Annual Meeting of the Association for Computational Linguistics (Volume 1: Long Papers)},
  pages = {9313--9332},
  address = {Bangkok, Thailand},
  publisher = {Association for Computational Linguistics},
  year = {2024},
  doi = {10.18653/v1/2024.acl-long.505}
}

@article{glm2025,
  author = {{GLM-V Team}},
  title = {{GLM-4.5V} and {GLM-4.1V-Thinking}: Towards versatile multimodal reasoning with scalable reinforcement learning},
  journal = {arXiv preprint arXiv:2507.01006},
  year = {2025},
  doi = {10.48550/arXiv.2507.01006}
}

@inproceedings{huang2024selfcorrect,
  author = {Jie Huang and Xinyun Chen and Swaroop Mishra and Huaixiu Steven Zheng and Adams Wei Yu and Xinying Song and Denny Zhou},
  title = {Large language models cannot self-correct reasoning yet},
  booktitle = {The Twelfth International Conference on Learning Representations},
  year = {2024}
}

@inproceedings{li2021prefix,
  author = {Xiang Lisa Li and Percy Liang},
  title = {Prefix-tuning: Optimizing continuous prompts for generation},
  booktitle = {Proceedings of the 59th Annual Meeting of the Association for Computational Linguistics and the 11th International Joint Conference on Natural Language Processing (Volume 1: Long Papers)},
  pages = {4582--4597},
  address = {Online},
  publisher = {Association for Computational Linguistics},
  year = {2021},
  doi = {10.18653/v1/2021.acl-long.353}
}

@inproceedings{li2024androidcontrol,
  author = {Wei Li and William Bishop and Alice Li and Chris Rawles and Folawiyo Campbell-Ajala and Divya Tyamagundlu and Oriana Riva},
  title = {On the effects of data scale on {UI} control agents},
  booktitle = {Advances in Neural Information Processing Systems (Datasets and Benchmarks Track)},
  volume = {37},
  pages = {92130--92154},
  year = {2024}
}

@inproceedings{lu2024guiodyssey,
  author = {Quanfeng Lu and Wenqi Shao and Zitao Liu and Lingxiao Du and Fanqing Meng and Boxuan Li and Botong Chen and Siyuan Huang and Kaipeng Zhang and Ping Luo},
  title = {{GUIOdyssey}: A comprehensive dataset for cross-app {GUI} navigation on mobile devices},
  booktitle = {Proceedings of the IEEE/CVF International Conference on Computer Vision (ICCV)},
  pages = {22404--22414},
  year = {2025},
  url = {https://openaccess.thecvf.com/content/ICCV2025/html/Lu_GUIOdyssey_A_Comprehensive_Dataset_for_Cross-App_GUI_Navigation_on_Mobile_ICCV_2025_paper.html}
}

@inproceedings{madaan2023selfrefine,
  author = {Aman Madaan and Niket Tandon and Prakhar Gupta and Skyler Hallinan and Luyu Gao and Sarah Wiegreffe and Uri Alon and Nouha Dziri and Shrimai Prabhumoye and Yiming Yang and Shashank Gupta and Bodhisattwa Prasad Majumder and Katherine Hermann and Sean Welleck and Amir Yazdanbakhsh and Peter Clark},
  title = {{Self-Refine}: Iterative refinement with self-feedback},
  booktitle = {Advances in Neural Information Processing Systems},
  volume = {36},
  pages = {46534--46594},
  year = {2023}
}

@article{qin2025uitars,
  author = {Yujia Qin and Yining Ye and Junjie Fang and Haoming Wang and Shihao Liang and Shizuo Tian and Junda Zhang and Jiahao Li and Yunxin Li and Shijue Huang and Wanjun Zhong and Kuanye Li and Jiale Yang and Yu Miao and Woyu Lin and Longxiang Liu and Xu Jiang and Qianli Ma and Jingyu Li and Xiaojun Xiao and Kai Cai and Chuang Li and Yaowei Zheng and Chaolin Jin and Chen Li and Xiao Zhou and Minchao Wang and Haoli Chen and Zhaojian Li and Haihua Yang and Haifeng Liu and Feng Lin and Tao Peng and Xin Liu and Guang Shi},
  title = {{UI-TARS}: Pioneering automated {GUI} interaction with native agents},
  journal = {arXiv preprint arXiv:2501.12326},
  year = {2025},
  doi = {10.48550/arXiv.2501.12326}
}

@inproceedings{rawles2023aitw,
  author = {Christopher Rawles and Alice Li and Daniel Rodriguez and Oriana Riva and Timothy Lillicrap},
  title = {{Android in the Wild}: A large-scale dataset for {Android} device control},
  booktitle = {Advances in Neural Information Processing Systems (Datasets and Benchmarks Track)},
  volume = {36},
  pages = {59708--59728},
  year = {2023}
}

@inproceedings{rawles2025androidworld,
  author = {Christopher Rawles and Sarah Clinckemaillie and Yifan Chang and Jonathan Waltz and Gabrielle Lau and Marybeth Fair and Alice Li and William Bishop and Wei Li and Folawiyo Campbell-Ajala and Daniel Toyama and Robert Berry and Divya Tyamagundlu and Timothy Lillicrap and Oriana Riva},
  title = {{AndroidWorld}: A dynamic benchmarking environment for autonomous agents},
  booktitle = {The Thirteenth International Conference on Learning Representations},
  year = {2025}
}

@article{venus2026,
  author = {{Venus Team} and Changlong Gao and Zhangxuan Gu and Yulin Liu and Xinyu Qiu and Shuheng Shen and Yue Wen and Tianyu Xia and Zhenyu Xu and Zhengwen Zeng and Beitong Zhou and Xingran Zhou and Weizhi Chen and Sunhao Dai and Jingya Dou and Yichen Gong and Yuan Guo and Zhenlin Guo and Feng Li and Qian Li and Jinzhen Lin and Yuqi Zhou and Linchao Zhu and Liang Chen and Zhenyu Guo and Changhua Meng and Weiqiang Wang},
  title = {{UI-Venus-1.5} technical report},
  journal = {arXiv preprint arXiv:2602.09082},
  year = {2026},
  doi = {10.48550/arXiv.2602.09082}
}

@inproceedings{wang2024mobileagentv2,
  author = {Junyang Wang and Haiyang Xu and Haitao Jia and Xi Zhang and Ming Yan and Weizhou Shen and Ji Zhang and Fei Huang and Jitao Sang},
  title = {{Mobile-Agent-v2}: Mobile device operation assistant with effective navigation via multi-agent collaboration},
  booktitle = {Advances in Neural Information Processing Systems},
  volume = {37},
  year = {2024}
}

@inproceedings{wang2025awm,
  author = {Zora Zhiruo Wang and Jiayuan Mao and Daniel Fried and Graham Neubig},
  title = {Agent workflow memory},
  booktitle = {Proceedings of the 42nd International Conference on Machine Learning},
  series = {Proceedings of Machine Learning Research},
  volume = {267},
  year = {2025}
}

@article{wang2025internvl,
  author = {Weiyun Wang and Zhangwei Gao and Lixin Gu and Hengjun Pu and Long Cui and Xingguang Wei and Zhaoyang Liu and Linglin Jing and Shenglong Ye and Jie Shao and Zhaokai Wang and Zhe Chen and Hongjie Zhang and Ganlin Yang and Haomin Wang and Qi Wei and Jinhui Yin and Wenhao Li and Erfei Cui and Guanzhou Chen and Zichen Ding and Changyao Tian and Zhenyu Wu and Jingjing Xie and Zehao Li and Bowen Yang and Yuchen Duan and Xuehui Wang and Zhi Hou and Haoran Hao and Tianyi Zhang and Songze Li and Xiangyu Zhao and Haodong Duan and Nianchen Deng and Bin Fu and Yinan He and Yi Wang and Conghui He and Botian Shi and Junjun He and Yingtong Xiong and Han Lv and Lijun Wu and Wenqi Shao and Kaipeng Zhang and Huipeng Deng and Biqing Qi and Jiaye Ge and Qipeng Guo and Wenwei Zhang and Songyang Zhang and Maosong Cao and Junyao Lin and Kexian Tang and Jianfei Gao and Haian Huang and Yuzhe Gu and Chengqi Lyu and Huanze Tang and Rui Wang and Haijun Lv and Wanli Ouyang and Limin Wang and Min Dou and Xizhou Zhu and Tong Lu and Dahua Lin and Jifeng Dai and Weijie Su and Bowen Zhou and Kai Chen and Yu Qiao and Wenhai Wang and Gen Luo},
  title = {{InternVL3.5}: Advancing open-source multimodal models in versatility, reasoning, and efficiency},
  journal = {arXiv preprint arXiv:2508.18265},
  year = {2025},
  doi = {10.48550/arXiv.2508.18265}
}

@article{wang2025mobileagente,
  author = {Zhenhailong Wang and Haiyang Xu and Junyang Wang and Xi Zhang and Ming Yan and Ji Zhang and Fei Huang and Heng Ji},
  title = {{Mobile-Agent-E}: Self-evolving mobile assistant for complex tasks},
  journal = {arXiv preprint arXiv:2501.11733},
  year = {2025},
  doi = {10.48550/arXiv.2501.11733}
}

@article{wu2025comem,
  author = {Wenyi Wu and Kun Zhou and Ruoxin Yuan and Vivian Yu and Stephen Wang and Zhiting Hu and Biwei Huang},
  title = {Auto-scaling continuous memory for {GUI} agent},
  journal = {arXiv preprint arXiv:2510.09038},
  year = {2025},
  doi = {10.48550/arXiv.2510.09038}
}

@inproceedings{wu2025osatlas,
  author = {Zhiyong Wu and Zhenyu Wu and Fangzhi Xu and Yian Wang and Qiushi Sun and Chengyou Jia and Kanzhi Cheng and Zichen Ding and Liheng Chen and Paul Pu Liang and Yu Qiao},
  title = {{OS-ATLAS}: Foundation action model for generalist {GUI} agents},
  booktitle = {The Thirteenth International Conference on Learning Representations},
  year = {2025}
}

@article{wu2025resum,
  author = {Xixi Wu and Kuan Li and Yida Zhao and Liwen Zhang and Litu Ou and Huifeng Yin and Zhongwang Zhang and Xinmiao Yu and Dingchu Zhang and Yong Jiang and Pengjun Xie and Fei Huang and Minhao Cheng and Shuai Wang and Hong Cheng and Jingren Zhou},
  title = {{ReSum}: Unlocking long-horizon search intelligence via context summarization},
  journal = {arXiv preprint arXiv:2509.13313},
  year = {2025},
  doi = {10.48550/arXiv.2509.13313}
}

@inproceedings{xu2024aguvis,
  author = {Yiheng Xu and Zekun Wang and Junli Wang and Dunjie Lu and Tianbao Xie and Amrita Saha and Doyen Sahoo and Tao Yu and Caiming Xiong},
  title = {{Aguvis}: Unified pure vision agents for autonomous {GUI} interaction},
  booktitle = {Proceedings of the 42nd International Conference on Machine Learning},
  series = {Proceedings of Machine Learning Research},
  volume = {267},
  pages = {69772--69805},
  publisher = {PMLR},
  year = {2025}
}

@article{xu2026mobileagent,
  author = {Haiyang Xu and Xi Zhang and Haowei Liu and Junyang Wang and Zhaozai Zhu and Shengjie Zhou and Xuhao Hu and Feiyu Gao and Junjie Cao and Zihua Wang and Zhiyuan Chen and Jitong Liao and Qi Zheng and Jiahui Zeng and Ze Xu and Shuai Bai and Junyang Lin and Jingren Zhou and Ming Yan},
  title = {{Mobile-Agent-v3.5}: Multi-platform fundamental {GUI} agents},
  journal = {arXiv preprint arXiv:2602.16855},
  year = {2026},
  doi = {10.48550/arXiv.2602.16855}
}

@inproceedings{yan2021pct,
  author = {Sijie Yan and Yuanjun Xiong and Kaustav Kundu and Shuo Yang and Siqi Deng and Meng Wang and Wei Xia and Stefano Soatto},
  title = {Positive-congruent training: Towards regression-free model updates},
  booktitle = {Proceedings of the IEEE/CVF Conference on Computer Vision and Pattern Recognition (CVPR)},
  pages = {14299--14308},
  year = {2021},
  url = {https://openaccess.thecvf.com/content/CVPR2021/html/Yan_Positive-Congruent_Training_Towards_Regression-Free_Model_Updates_CVPR_2021_paper.html}
}

@article{yang2026guilibra,
  author = {Rui Yang and Qianhui Wu and Zhaoyang Wang and Hanyang Chen and Ke Yang and Hao Cheng and Huaxiu Yao and Baolin Peng and Huan Zhang and Jianfeng Gao and Tong Zhang},
  title = {{GUI-Libra}: Training native {GUI} agents to reason and act with action-aware supervision and partially verifiable {RL}},
  journal = {arXiv preprint arXiv:2602.22190},
  year = {2026},
  doi = {10.48550/arXiv.2602.22190}
}

@inproceedings{yao2023react,
  author = {Shunyu Yao and Jeffrey Zhao and Dian Yu and Nan Du and Izhak Shafran and Karthik Narasimhan and Yuan Cao},
  title = {{ReAct}: Synergizing reasoning and acting in language models},
  booktitle = {The Eleventh International Conference on Learning Representations},
  year = {2023}
}

@article{zeng2026memento,
  author = {Ziyun Zeng and Hang Hua and Bocheng Zou and Mu Cai and Rogerio Feris and Jiebo Luo},
  title = {{MementoGUI}: Learning agentic multimodal memory control for long-horizon {GUI} agents},
  journal = {arXiv preprint arXiv:2605.18652},
  year = {2026},
  doi = {10.48550/arXiv.2605.18652}
}

@inproceedings{zhang2025appagent,
  author = {Chi Zhang and Zhao Yang and Jiaxuan Liu and Yanda Li and Yucheng Han and Xin Chen and Zebiao Huang and Bin Fu and Gang Yu},
  title = {{AppAgent}: Multimodal agents as smartphone users},
  booktitle = {Proceedings of the 2025 CHI Conference on Human Factors in Computing Systems},
  year = {2025}
}

@article{zhang2026focusmem,
  author = {Zhuoran Zhang and Bowen Li and Jingcheng Ju and Yang Shi and Qixun Wang and Haotian Wang and Wei Chen and Tengjiao Wang},
  title = {{FocusMem}: Factorizing content, readout, and trust in latent {GUI} memory},
  journal = {arXiv preprint arXiv:2608.04530},
  year = {2026},
  doi = {10.48550/arXiv.2608.04530}
}

@article{zhang2026memw,
  author = {Guibin Zhang and Yaohui Ling and Fanci Meng and Kun Wang and Shuicheng Yan},
  title = {{Mem-W}: Latent memory-native {GUI} agents},
  journal = {arXiv preprint arXiv:2605.09317},
  year = {2026},
  doi = {10.48550/arXiv.2605.09317}
}

@inproceedings{zheng2024synapse,
  author = {Longtao Zheng and Rundong Wang and Xinrun Wang and Bo An},
  title = {Synapse: Trajectory-as-exemplar prompting with memory for computer control},
  booktitle = {The Twelfth International Conference on Learning Representations},
  year = {2024}
}

@article{zheng2026taskstate,
  author = {Yujie Zheng and Zikang Liu and Xin Zhao and Ji-Rong Wen},
  title = {A task-state representation for long-horizon mobile {GUI} agents},
  journal = {arXiv preprint arXiv:2607.00502},
  year = {2026},
  doi = {10.48550/arXiv.2607.00502}
}

@inproceedings{zhu2026hymem,
  author = {Sibo Zhu and Wenyi Wu and Kun Zhou and Stephen Wang and Biwei Huang},
  title = {Hybrid self-evolving structured memory for computer-use agents},
  booktitle = {Findings of the Association for Computational Linguistics: ACL 2026},
  pages = {11287--11304},
  address = {San Diego, California, United States},
  publisher = {Association for Computational Linguistics},
  year = {2026},
  url = {https://aclanthology.org/2026.findings-acl.549/}
}

@inproceedings{welleck2023selfcorrect,
  author = {Sean Welleck and Ximing Lu and Peter West and Faeze Brahman and Tianxiao Shen and Daniel Khashabi and Yejin Choi},
  title = {Generating sequences by learning to self-correct},
  booktitle = {The Eleventh International Conference on Learning Representations},
  year = {2023}
}

@article{kamoi2024selfcorrect,
  author = {Ryo Kamoi and Yusen Zhang and Nan Zhang and Jiawei Han and Rui Zhang},
  title = {When can {LLMs} actually correct their own mistakes? {A} critical survey of self-correction of {LLMs}},
  journal = {Transactions of the Association for Computational Linguistics},
  volume = {12},
  pages = {1417--1440},
  year = {2024},
  doi = {10.1162/tacl_a_00713}
}

@inproceedings{zhu2026helamem,
  title     = {{HeLa-Mem}: Hebbian Learning and Associative Memory for {LLM} Agents},
  author    = {Zhu, Jinchang and Li, Jindong and Zhang, Cheng and Liu, Jiahong and Yang, Menglin},
  booktitle = {Proceedings of the 64th Annual Meeting of the Association for Computational Linguistics (Volume 1: Long Papers)},
  pages     = {13757--13769},
  address   = {San Diego, California, United States},
  publisher = {Association for Computational Linguistics},
  year      = {2026}
}

@article{zhang2026memcut,
  author = {Cheng Zhang and Xuanning Cao and Zhicheng Zheng and Hui Xiong},
  title = {{MemCut}: Every secret leaves a trail in the long-term memory of {LLM} agents},
  journal = {arXiv preprint arXiv:2610.21405},
  year = {2026},
  doi = {10.48550/arXiv.2610.21405},
  url = {https://arxiv.org/abs/2610.21405}
}

\appendix
\setlength{\emergencystretch}{1.5em}
\setcounter{topnumber}{5}
\setcounter{bottomnumber}{2}
\setcounter{totalnumber}{7}
\setcounter{dbltopnumber}{3}
\renewcommand{\topfraction}{0.95}
\renewcommand{\bottomfraction}{0.5}
\renewcommand{\textfraction}{0.05}
\renewcommand{\floatpagefraction}{0.8}
\renewcommand{\dbltopfraction}{0.95}
\renewcommand{\dblfloatpagefraction}{0.8}
\setlength{\floatsep}{9pt plus 2pt minus 2pt}
\setlength{\textfloatsep}{12pt plus 2pt minus 3pt}
\setlength{\dblfloatsep}{9pt plus 2pt minus 2pt}
\setlength{\dbltextfloatsep}{12pt plus 2pt minus 3pt}
\makeatletter %
\setlength{\@fptop}{0pt}\setlength{\@fpsep}{9pt plus 2pt}\setlength{\@fpbot}{0pt plus 1fil}
\setlength{\@dblfptop}{0pt}\setlength{\@dblfpsep}{9pt plus 2pt}\setlength{\@dblfpbot}{0pt plus 1fil}
\makeatother

\section{Evidence for Scoped Revision}
\label{app:ablation}

The ablation set contains 1,463 held-out steps: 731 from 128 AndroidControl trajectories and 732 from 50 GUI-Odyssey trajectories. These trajectories do not overlap the main test sets. The plain actor gets 879 steps right and 584 wrong. Paired 95\% intervals use 20,000 dataset-stratified whole-trajectory bootstrap resamples. Results in this section use the main run; Appendix~\ref{app:seeds} reports repeated runs. CE denotes cross-entropy alone, and scoped and unscoped denote preservation on correct first opinions and on all steps, respectively.

\subsection{The objective and the first opinion}

Table~\ref{tab:contrasts} gives the key paired comparisons behind \S\ref{sec:analysis}. Scoped preservation improves the first-opinion encoder by 1.98\,pp over CE, and by 2.73\,pp over unscoped preservation. Its advantage over a scoped task-only encoder and its main-run interaction with the first opinion remain uncertain: both intervals include zero.

\begin{table}[t]
\centering
\tablefont
\setlength{\tabcolsep}{3pt}
\adjustbox{max width=\columnwidth}{%
\begin{tabular}{Lrrc}
\toprule
\hd{Contrast} & Steps & pp & 95\% CI\\
\midrule
\firstband{4}{Scoped minus cross-entropy alone}
task-only encoder & 9 & +0.62 & [$-$0.74, 1.89]\\
history encoder & 10 & +0.68 & [$-$0.59, 1.94]\\
first-opinion encoder & 29 & +1.98 & [0.53, 3.44]\\
interaction & 19 & +1.30 & [$-$0.38, 3.02]\\
\band{4}{\textsc{ScopeMem} minus}
first opinion, unscoped & 40 & +2.73 & [0.98, 4.55]\\
first epoch only & 31 & +2.12 & [0.85, 3.40]\\
task only, scoped & 21 & +1.44 & [$-$0.49, 3.56]\\
fixed-query & 124 & +8.48 & [6.46, 10.56]\\
\band{4}{Between prefixes}
first opinion $-$ history, scoped & 23 & +1.57 & [$-$0.25, 3.48]\\
first opinion $-$ history, CE & 4 & +0.27 & [$-$1.50, 2.05]\\
task only $-$ history, CE & 3 & +0.21 & [$-$1.62, 1.91]\\
\bottomrule
\end{tabular}}
\caption{Paired contrasts on the ablation set: difference in correct steps and in accuracy (pp), with 95\% trajectory-bootstrap CIs. Interaction: scoping effect of the first-opinion encoder minus that of the history encoder.}
\label{tab:contrasts}
\end{table}

Table~\ref{tab:decomp} separates how often a prefix changes an action from how often that change succeeds. Scoping reduces changes to correct actions for all three encoders. For the first-opinion encoder, it retains the rate of changes to errors and improves their repair yield; the task-only and history encoders instead reduce changes to errors as well.

\begin{table}[t]
\centering
\tablefont
\setlength{\tabcolsep}{3pt}
\adjustbox{max width=\columnwidth}{%
\begin{tabular}{lrrrr}
\toprule
 & \multicolumn{2}{c}{\gpill{Errors}} & \multicolumn{2}{c}{\gpill{Correct actions}}\\
\gcmidd{2-3}{4-5}
Variant & changed & repaired & changed & harmed\\
\midrule
Task only, CE & 90.58 & 46.50 & 71.90 & 11.23\\
Task only, scoped & 83.90 & 49.39 & 63.71 & 10.36\\
History, CE & 90.07 & 45.63 & 71.44 & 10.83\\
History, scoped & 84.25 & 47.76 & 64.05 & 9.41\\
First opinion, CE & 89.55 & 46.46 & 71.67 & 10.63\\
First opinion, unscoped & 80.14 & 45.30 & 58.59 & 9.13\\
\hl \textsc{ScopeMem} & 89.73 & 48.47 & 62.46 & 8.93\\
\bottomrule
\end{tabular}}
\caption{Share (\%) of the plain actor's errors and correct actions that each prefix changes, and of those changes that repair or harm. Coordinate shifts within tolerance count as changes.}
\label{tab:decomp}
\end{table}

\subsection{First-opinion interventions}
\label{app:prefix}

Table~\ref{tab:prefix} tests whether the encoder uses the first opinion. Perturbing it or replacing it with one from another step lowers accuracy. These interventions keep the encoder fixed; removing history uses a separately trained encoder. The history-ablation interval includes zero.

\begin{table}[t]
\centering
\tablefont
\setlength{\tabcolsep}{3.4pt}
\adjustbox{max width=\columnwidth}{%
\begin{tabular}{Lrrrc}
\toprule
\hd{Condition} & R/H & Gain & $\Delta$ & 95\% CI\\
\midrule
\hl \oursname & 254/49 & 14.01 & -- & --\\
\band{5}{Alter the first opinion}
perturbed & 248/58 & 12.99 & $-$1.03 & [$-$1.90, $-$0.21]\\
taken from another step & 244/63 & 12.37 & $-$1.64 & [$-$2.74, $-$0.66]\\
\band{5}{Remove an encoder input}
history & 248/56 & 13.12 & $-$0.89 & [$-$2.40, 0.60]\\
\bottomrule
\end{tabular}}
\caption{Interventions on \textsc{ScopeMem}: repairs/harms against the plain actor, gain (pp), and difference $\Delta$ from \textsc{ScopeMem} (pp) with its paired 95\% CI.}
\label{tab:prefix}
\end{table}

\subsection{Which errors are avoided?}
\label{app:harms}

Relative to the first-opinion encoder trained with CE, \textsc{ScopeMem} reduces action-type harms from 35 to 19, while coordinate harms change from 22 to 21. Other-argument harms change from 10 to 8, and parse harms from 0 to 1, giving totals of 67 and 49. Thus, most of the reduction concerns the decision tokens that receive stronger preservation.

Completion illustrates why preservation must be scoped. Of 48 reference COMPLETE actions, the CE variant finds 37 with 8 premature completions. Unscoped preservation finds only 30, with 2 premature completions. \textsc{ScopeMem} finds 38 with the same 2 premature completions, preserving the reduction in false completions without losing recall.

\section{Comparison with Mem-W}
\label{app:memw}

The comparisons use sampled test trajectories and the shared actor context; general-purpose and GUI-specialised actors do not use their native history formats. In Table~\ref{tab:memw}, ``only one keeps'' counts correct plain-actor decisions preserved by only one method, while ``only one repairs'' counts errors repaired by only one method. The differences sum to the accuracy advantage. AITW counts pool all 836 steps, so the 2.03\,pp advantage differs from the macro-average gap in Table~\ref{tab:main}. Intervals resample whole trajectories, within subsets for AITW.

\paragraph{Actions and completion.} Clicks and long-presses account for 48 of the 80 additional correct steps across AndroidControl and GUI-Odyssey. On GUI-Odyssey, completion decisions account for 24 of the 69-step advantage, partly overlapping the click category. Among its 96 reference COMPLETE actions, Mem-W finds 61 with 15 premature completions; \textsc{ScopeMem} finds 75 with 6.

\paragraph{Trajectory length.} Splitting GUI-Odyssey at its median of 14 steps gives 581 steps in 52 short trajectories and 1,028 in 48 long ones. \textsc{ScopeMem}'s advantage over Mem-W is similar in both groups: 4.48\,pp [1.68, 7.56] and 4.18\,pp [2.44, 5.91], respectively. This does not suggest that its advantage comes mainly from longer histories.

\section{Consistency across Training Runs}
\label{app:seeds}

We repeat encoder training three times (seeds 17, 1031 and 2027). Table~\ref{tab:seedmain} reports the main benchmark results; Table~\ref{tab:seedgain} shows the repair--harm trade-off for the key ablations. Scoped preservation lowers harms for every encoder in every run. The first-opinion encoder has the largest accuracy benefit from scoping in each run (Table~\ref{tab:seedgain}). Its interaction with scoping averages 1.18\,pp [0.38, 1.99]. These 95\% $t$ intervals describe variation across three training runs, not test-set sampling uncertainty.

\begin{table}[t]
\centering
\tablefont
\setlength{\tabcolsep}{4pt}
\adjustbox{max width=\columnwidth}{%
\begin{tabular}{Lccc}
\toprule
\hd{Run} & AndroidControl & GUI-Odyssey & AITW\\
\midrule
main & 67.91 & 72.90 & 66.69\\
second & 67.20 & 72.47 & 66.11\\
third & 68.44 & 73.28 & 67.04\\
\midrule
Mean $\pm$ SD & 67.85 $\pm$ 0.62 & 72.88 $\pm$ 0.40 & 66.61 $\pm$ 0.47\\
95\% interval & [66.30, 69.40] & [71.88, 73.89] & [65.45, 67.78]\\
\bottomrule
\end{tabular}}
\caption{\textsc{ScopeMem} on the main test sets in each run (EM; macro accuracy for AITW), with mean, SD and 95\% $t$ interval across runs.}
\label{tab:seedmain}
\end{table}

\begin{table}[t]
\centering
\tablefont
\setlength{\tabcolsep}{4pt}
\begin{tabular}{lccc}
\toprule
Variant & Main & Second & Third\\
\midrule
Task only, CE & 246/71 & 250/70 & 241/70\\
Task only, scoped & 242/58 & 248/57 & 234/58\\
History, CE & 240/68 & 243/66 & 242/67\\
History, scoped & 235/53 & 236/50 & 234/54\\
First opinion, CE & 243/67 & 246/64 & 249/64\\
First opinion, unscoped & 212/47 & 217/45 & 217/48\\
\hl \textsc{ScopeMem} & 254/49 & 250/47 & 263/52\\
\bottomrule
\end{tabular}
\caption{Scoped preservation reduces harms for every encoder in all three runs. Repairs/harms on the ablation set in each training run.}
\label{tab:seedgain}
\end{table}

\end{document}